\documentclass{article}

     \usepackage[final]{neurips_2019}

\usepackage[utf8]{inputenc} 
\usepackage[T1]{fontenc}    
\usepackage{hyperref}       
\usepackage{url}            
\usepackage{booktabs}       
\usepackage{amsfonts}       
\usepackage{nicefrac}       
\usepackage{microtype}      
\usepackage{graphicx}
\graphicspath{ {./images/} }

\title{Mind The Facts: Knowledge-Boosted Coherent Abstractive Text Summarization}

 \author{
   Beliz Gunel\\
   Department of Electrical Engineering\\
   Stanford University\\
   Stanford, CA 94305 \\
   \texttt{bgunel@stanford.edu} \\
   \And
   Chenguang Zhu, Michael Zeng, Xuedong Huang \\
   AI Cognitive Services Research Group\\
   Microsoft\\
   Redmond, WA 98052\\
   \texttt{\{chezhu, nzeng, xdh\}@microsoft.com} \\
  }

\begin{document}

\maketitle

\begin{abstract}
Neural models have become successful at producing abstractive summaries that are human-readable and fluent. However, these models have two critical shortcomings: they often don't respect the facts that are either included in the source article or are known to humans as commonsense knowledge, and they don't produce coherent summaries when the source article is long. In this work, we propose a novel architecture that extends Transformer [4] encoder-decoder architecture in order to improve on these shortcomings. First, we incorporate entity-level knowledge from the Wikidata [15] knowledge graph into the encoder-decoder architecture. Injecting structural world knowledge from Wikidata helps our abstractive summarization model to be more fact-aware. Second, we utilize the ideas used in Transformer-XL [8] language model in our proposed encoder-decoder architecture. This helps our model with producing coherent summaries even when the source article is long. We test our model on CNN/Daily Mail [6, 13] summarization dataset and show improvements on ROUGE [17] scores over the baseline Transformer model. We also include model predictions for which our model accurately conveys the facts, while the baseline Transformer model doesn't.
\end{abstract}

\section{Introduction}
Summarization is the task of generating a shorter text that contains the key information from source text, and the task is a good measure for natural language understanding and generation. Broadly, there are two approaches in summarization: extractive and abstractive. Extractive approaches simply select and rearrange sentences from the source text to form the summary. There has been many neural models proposed for extractive summarization over the past years [11, 18]. Current state-of-the-art model for the extractive approach fine-tunes a simple variant of the popular language model BERT [12] for the extractive summarization task [10].

On the other hand, abstractive approaches generate novel text, and are able to paraphrase sentences while forming the summary. This is a hard task even for humans, and it's hard to evaluate due to the subjectivity of what is considered a "ground truth" summary during evaluation. Recently, many neural abstractive summarization models have been proposed that use either LSTM-based sequence-to-sequence attentional models or Transformer as their backbone architectures [1, 3, 6, 9]. These models also integrate various techniques to their backbone architecture such as coverage, copy mechanism and content selector module in order to enhance their performance. There is also some recent work on abstractive summarization based on reinforcement learning techniques that optimize objectives in addition to the standard maximum likelihood loss [1, 2].

Although current neural abstractive summarization models can achieve high ROUGE scores on popular benchmarks and are able to produce fluent summaries, they have two main shortcomings: i. They don't respect the facts that are either included in the source article or are known to humans as commonsense knowledge; ii. They don't produce coherent summaries when the source article is long. 

In this work, we propose a novel architecture that extends 
Transformer encoder-decoder architecture to improve on these challenges. First, we incorporate entity-level knowledge from the Wikidata knowledge graph into the encoder-decoder architecture. Injecting structural world knowledge from Wikidata helps our abstractive summarization model to be more fact-aware. Second, we utilize the ideas used in Transformer-XL language model in our encoder-decoder architecture. This helps our model with producing coherent summaries even when the source article is long.

\section{Proposed Method}
\label{headings}

\subsection{Transformer vs. Transformer-XL}
Recently, Transformer architectures have been immensely successful in various natural language processing applications including neural machine translation, question answering and neural summarization and pretrained language modeling. However, Transformers have fixed-length context, which results in worse performance while encoding long source text. In addition, these fixed-length context segments do not respect the sentence boundaries, resulting in context fragmentation which is a problem even for the short sequences. Recently, Transformer-XL has offered an effective solution for this long-range dependency problem in the context of language modeling. They have introduced the notion of recurrence into a self-attention-based model by reusing hidden states from the previous segments, and have introduced the idea of relative positional encoding to make the recurrence scheme possible. Transformer-XL has state-of-the-art perplexity performance, learns dependency 450\% longer than vanilla Transformers, and is up to 1,800+ times faster than vanilla Transformers at inference time on language modeling tasks. 

Inspired by the strong performance of the Transformer-XL language model on modeling long-range dependency, we extend Transformer-XL to an encoder-decoder architecture based on the Transformer architecture. In other words, we calculate the attention scores at every multi-head attention layer in our architecture shown in Figure 1 based on Transformer-XL attention decomposition. We compare the attention decompositions of vanilla Transformer and Transformer-XL. Below equations show the attention computation between query $q_i$ and key vector $k_j$ within the same segment. U matrix shows the absolute positional encoding, E matrix is the token embedding matrix, $W_q$ and $W_k$ represent the query and key matrices. In the Transformer-XL attention formulation, $R_{i-j}$ is the relative positional encoding matrix without trainable parameters, and u, v, $W_{k, R}$, $W_{k,E}$ are all trainable parameters. 
\[A^{vanilla}_{i,j} = E^{T}_{x_i}W^{T}_qW_kE_{x_j}+E^{T}_{x_i}W^{T}_qW_kU_{j}+U_i^{T}W^{T}_qW_kE_{x_j}+U_i^{T}W^{T}_qW_kU_j \] 
\[A^{xl}_{i,j} = E^{T}_{x_i}W^{T}_qW_{k,E}E_{x_j}+E^{T}_{x_i}W^{T}_qW_{k,R}R_{i-j}+u^{T}W_{k,E}E_{x_j}+v^{T}W_{k,R}R_{i-j} \]

Overall, Transformer-XL's architecture is shown below for a segment $\tau$ in the $n$-th transformer layer. $SG$ denotes stop-gradient, and $\circ$ denotes concatenation. We refer the readers to the original Transformer-XL paper [8] for further discussion on the new parameterization for attention calculations and more details on the design decisions for the architecture. \\

\[\tilde{h}^{n-1}_{\tau} = [SG(h^{n-1}_{\tau-1}) \circ h^{n-1}_{\tau}]\]\[q^n_{\tau}, k^n_{\tau}, v^n_{\tau} = h^{n-1}_{\tau}{W^n_q}^T, \tilde{h}^{n-1}_{\tau}{W^n_{k,E}}^T,  \tilde{h}^{n-1}_{\tau}{W^n_v}^T\]
\[A^n_{\tau, i, j} = {q^n_{\tau,i}}^Tk^n_{\tau,j}+{q^n_{\tau,i}}^TW^n_{k,R}R_{i-j}+u^Tk^n_{\tau,j}+v^{T}W^n_{k,R}R_{i-j}\] \\

It is important to note that, as in the vanilla Transformer, we still have the fully connected feed-forward network layers after multi-head attention layers, and residual connections around sublayers followed by layer normalizations. These layers are omitted in Figure 1 for simplicity. Empirically, we observe much more coherent articles with Transformer-XL encoder-decoder architecture compared to the Transformer baseline. Figure 3 shows a comparison for an input source article sampled from CNN/Daily Mail dataset.

\subsection{Wikidata Knowledge Graph Entity Embeddings}

Wikidata is a free and open multi-relational knowledge graph that serves as the central storage for the structured data of its many services including Wikipedia. We sample part of Wikidata that has 5 million entities and 25 million relationship triples. We learn entity embeddings for these sampled entities through the popular multi-relational data modeling method TransE [14]. 

TransE is a simple yet very powerful method that represents relationships between fact triples as translations operating in the low-dimensional entity embedding space. Specifically, we minimize a margin-based ranking criterion over the entity and relationship set using $L_2$ norm as the dissimilarity measure, d, as shown in the below equation. $S$ is the set of relationship triplets $(h, l, t)$ where $h$ and $t$ are entities in the set of entities $E$, and $l$ represents the relationships in the set of relationships $\mathcal{L}$. We construct the corrupted relationship triplets, which forms the "negative" set for the margin-based objective, through replacing either the head or tail of the relationship triple by a random entity. Low-dimensional entity and relationship embeddings are optimized through stochastic gradient descent with the constraint that $L_2$-norms of the entity embeddings are 1 (on the unit sphere), which is important in order to obtain meaningful embeddings.

\[\mathcal{L} = \sum_{(h, l, t) \in S}\sum_{(h', l, t') \in S'_{(h, l, t)}}[\gamma + d(h+l,t) - d(h'+l,t')]_{+}\] where $[x]_{+}$ denotes the positive part of $x$, $\gamma > 0$ is a margin hyperparameter, and \[S'_{(h, l, t)} = \{(h', l , t) | h' \in E \} \cup \{(h, l , t') | t' \in E \}\]

\subsection{Our Model Architecture}

Our overall model architecture is shown in Figure 1. We extend the encoder-decoder architecture such that the entity information can be effectively incorporated into the model. On the encoder side, we have a separate attention channel for the entities in parallel to the attention channel for the tokens. These two channels are followed by multi-head token self attention and multi-head cross token-entity attention. On the decoder side, we have multi-head masked token self-attention, multi-head masked entity self-attention, and multi-head cross attention between encoder and decoder, respectively. Finally, we have another layer of multi-head token attention followed by a feed-forward layer and softmax to output the tokens. Multi-head attention are conducted based on the Transformer-XL decomposition as in Section 2.1. 

Entity Linker modules use an off-the-shelf entity extractor and disambiguate the extracted entities to the Wikidata knowledge graph. Extracted entities are initialized with the pretrained Wikidata knowledge graph entity embeddings that are learned through TransE, as discussed in Section 2.2. Entity Conversion Learner modules use a series of feed-forward layers with ReLU activation. These modules learn entities that are in the same subspace with the corresponding tokens in the text.

\begin{figure}[htbp]
  \centering
  \includegraphics[scale=0.3]{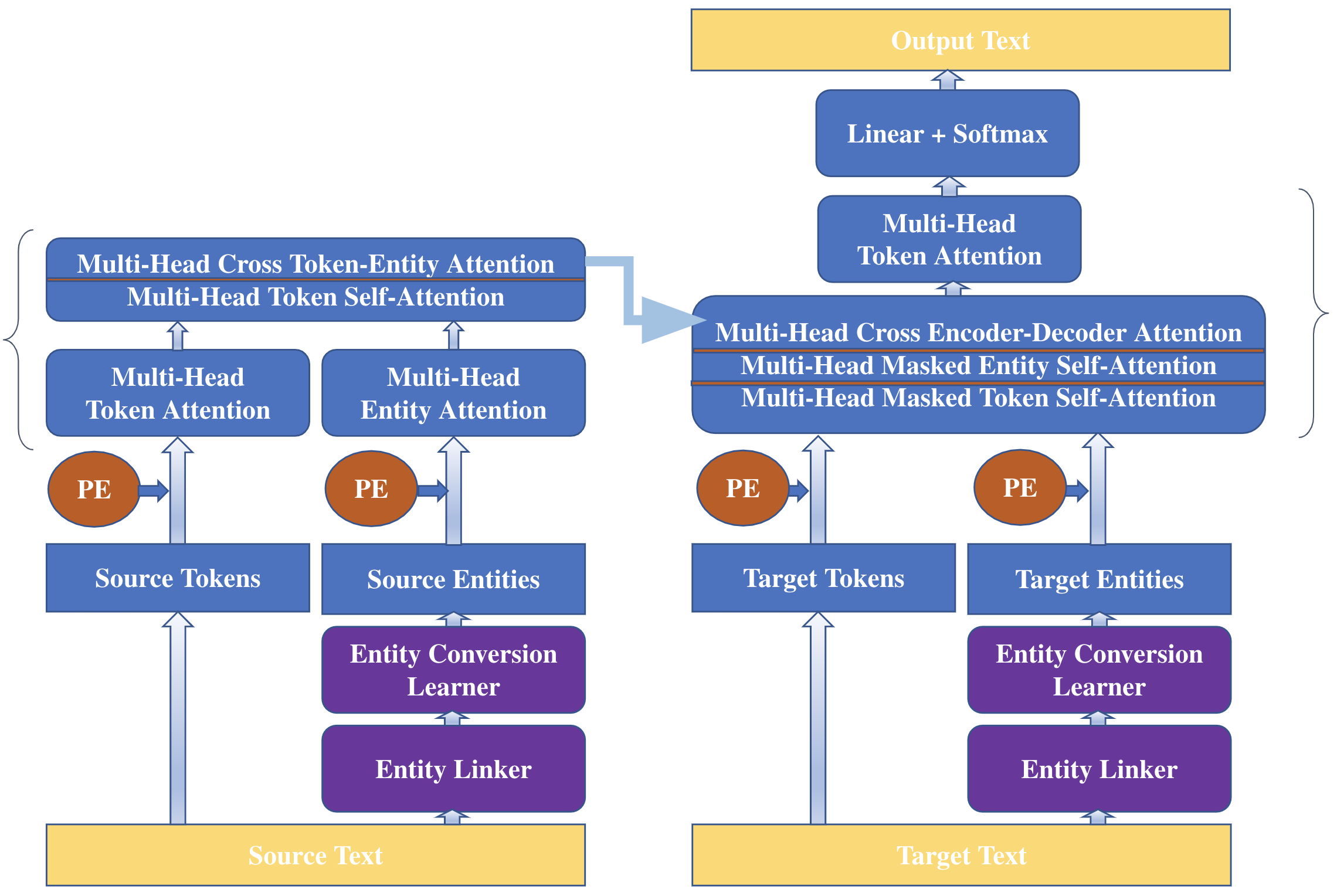}
  \caption{Our model architecture. PE stands for positional encoding. Single encoder and decoder layers are shown in parenthesis. In multi-layer architectures, these layers in curly brackets are stacked.}
\end{figure}


\section{Experiments}

\subsection{Dataset}

We evaluate our models on the benchmark dataset for summarization, CNN/Daily Mail. The dataset contains online news articles (781 words on average) paired with multi-sentence summaries (56 words on average). We use the standard splits that include 287,226 training pairs, 13,368 validation pairs and 11,490 test pairs. We do not anonymize the entities, instead operate directly on the original text. We truncate the articles to 400 tokens, and summaries to 100 tokens in train time and 120 tokens in test time. During preprocessing, we do not remove the case for higher quality entity extraction in our entity linking module.

\subsection{Quantitative Results}

We evaluate our model and the baseline based on the ROUGE metric that compares the generated summary to the human-written ground truth summary and counts the overlap of 1-grams (ROUGE-1), 2-grams (ROUGE-2), and longest common sequence (ROUGE-L). We use the pyrouge [17] package to obtain our scores and report the F1 scores for all ROUGE types.

Our baseline is the vanilla Transformer encoder-decoder architecture that's commonly used as the backbone architecture in abstractive summarization models. For both the baseline and our proposed model, we use 2 transformer layers and 4 heads and utilize beam search for decoding. We use 300 dimensions for both entity and token embeddings, BERTAdam as the optimizer, and minimum sentence generation length of 60. After hyperparameter search, we set the learning rate to 0.00005, dropout rate to 0.3, beam width to 5 and maximum sentence length to 90 during inference. We start entity extraction at the decoder it produces 20 tokens.

Our results on CNN/Daily Mail dataset are shown in Table 1. Our model improves over the Transformer baseline by +0.45 ROUGE-1 points and +0.56 ROUGE-L points on the full test set. In fact, we see better improvements when we test our model on the higher entity density slice of the test set as demonstrated in Table 2. Specifically, our model improves over the baseline by +0.85 ROUGE-1 points and +1.08 ROUGE-L points on the test set article-summary pairs in which there are more than 50 entities in the source article. Also, we include results in Table 1 in which we initialized the entity embeddings randomly to test the benefit of using Wikidata KG entity embeddings. Using random entity embeddings decreased the model performance while using Wikidata KG entity embeddings increased the model performance both for vanilla Transformer and for Transformer-XL backbone architectures. This supports our hypothesis that injecting structural world knowledge from external knowledge bases to abstractive summarization models improves model performance.

\begin{table}[!h]
  \caption{Results on CNN/Daily Mail dataset. R used as an abbreviation for ROUGE.}
  \label{results-table}
  \centering
  \begin{tabular}{llll}
    \toprule
    \cmidrule(r){1-2}
    Model     & R-1     & R-2   & R-L \\
    \midrule
    Transformer Baseline & 33.351& 
12.473  & 30.663    \\
    Transformer-Entity  w/
Random Entity Emb
     & 33.047 & 11.536 & 30.487     \\
    Transformer-Entity w/
Wikidata KG Emb
     & 33.741  & 12.171 & 31.076  \\
\textbf{Transformer-XL-Entity w/ Wikidata KG Emb (Our Model)} & \textbf{33.804} & \textbf{12.509}
& \textbf{31.225}\\
    \bottomrule
  \end{tabular}
\end{table}

\begin{table}[!h]
  \caption{Results on CNN/Daily Mail dataset with high density entities. R used as an abbreviation for ROUGE. >50 ent denotes the slice of test data that has more than 50 entities in the source article.}
  \label{sample-table}
  \centering
  \begin{tabular}{llll}
    \toprule
    \cmidrule(r){1-2}
    Model     & R-1 (>50 ent) & R-2 (>50 ent) & R-L (>50 ent)\\
    \midrule
    Transformer Baseline & 33.423 & 12.46 & 30.97 \\
 
\textbf{Our model} & \textbf{34.273} & \textbf{13.018} & \textbf{32.048}\\
    \bottomrule
  \end{tabular}
\end{table}

\subsection{Qualitative Results}

We conduct qualitative analysis on our model's predicted summaries and include some samples here. In Figure 2, we compare the transformer baseline output to the output of our model for a sampled input article from CNN/Daily Mail corpus. Baseline model makes several factual errors based on our manual fact-checking: 1. Neither McClaren nor Paul Clement was 42 years old at the time when the article was published. 2. Neither Steve McClaren nor Paul Clement worked as a Manchester United boss. On the other hand, our model respects the facts through incorporating world knowledge from Wikidata knowledge graph. Again based on our manual fact checking, we find: 1. Paul Clement was indeed working in Real Madrid before he was appointed the manager of Derby County. 2. Although "England boss" is too broad, he did work at Chelsea 2009-2011.

In Figure 3, we compare the transformer baseline output to the output of the Transformer-XL encoder-decoder model without entity integration in order to test the effect of architecture on summary coherency. Baseline model produces an incoherent summary, while Transformer-XL encoder-decoder model outputs coherent, human readable summary.

\begin{figure}[!h]
  \centering
\noindent\fbox{%
    \parbox{\textwidth}{%
    \textbf{Ground Truth Summary}\\
    Steve McClaren is expected to take Newcastle job if Derby don't go up. Rams are currently battling for Championship promotion via the play-offs. Paul Clement is a leading candidate for job. Derby will make formal contact with Real Madrid if McClaren leaves.\\
    \textbf{Transformer Baseline Output}\\
    Steve McClaren is a leading candidate to replace Steve McClaren. The 42-year-old has established a reputation as one of European football's leading coaches in recent years, working on mainly under Carlo Ancelotti. The former Manchester United boss is keen to secure promotion into the Premier League next season.\\
    \textbf{Output of Our Model}\\
    Paul Clement is a leading candidate to replace Steve McClaren at Derby County. The former England boss has established a reputation as one of Europe's leading football coaches in recent years. Clement is currently a Real Madrid coach.
    }%
}
  \caption{Comparison of the transformer baseline output and the output of our proposed model. Ground truth summary is sampled from the CNN/Daily Mail summarization corpus. Baseline model makes factual errors, while our model respects the facts through incorporating entity-level knowledge from Wikidata knowledge graph.}
\end{figure}

\begin{figure}[!h]
  \centering
\noindent\fbox{%
    \parbox{\textwidth}{%
    \textbf{Transformer Baseline Output}\\
    Wayne Oliveira has scored four goals in seven games and Oliveira. Oliveira has recovered from training-ground ankle injury. Oliveira says he is "not happy he is injured but if it gives me an chance". Gomis has been ruled out for between three to four weeks after being injured. Oliveira believes Bafetimbi Gomis' form has made seven of his eight Swansea appearances. ...\\
    \textbf{Transformer-XL Output}\\
    The Portugal striker has been ruled out for between three to four weeks. Nelson Oliveira has been sidelined for four weeks with injury. He has scored four goals in seven matches and has recovered from a training-ground injury. The 23-year-old has made his Swansea debut in the 5-0 home defeat against Chelsea in January ...
    }%
}
  \caption{Comparison of the transformer baseline output and the output of Transformer-XL encoder-decoder model output. Source article is sampled from the CNN/Daily Mail summarization corpus. Baseline model produces an incoherent summary, while Transformer-XL encoder-decoder model outputs coherent, human readable summary.}
\end{figure}

\newpage
\section{Discussion and Future Work}

We present an end-to-end novel encoder-decoder architecture that effectively integrates entity-level knowledge from the Wikidata knowledge graph in the attention calculations and utilizes Tranformer-XL ideas to encode longer term dependency. We show performance improvements over a Transformer baseline under same resources (in terms of number of layers, number of heads, number of dimensions of hidden states, etc.) on the popular CNN/Daily Mail summarization dataset. We also conduct preliminary fact-checking and include examples for which our model is respectful to the facts while baseline Transformer model isn't. 
Similar to the previous works in abstractive summarization, we find that ROUGE metric is not representative of the performance in terms of human readability, coherence and factual correctness. ROUGE, by definition, rewards extractive strategies by evaluating based on word overlap between ground truth summary and output model summary. Metric is not flexible towards rephrasing, which limits model's ability to output abstractive summaries. It's also important to note that "ground truth" is subjective in the abstractive summarization setting, since there can be more than one correct abstractive summary to a source article. We believe finding metrics that are representative of the desired performance is an important research direction. Finally, we believe entity linking should be part of the end-to-end training instead of a separate pipeline in the beginning. It's possible that we lose valuable information both during entity extraction part and during disambiguation part to the chosen knowledge graph.

\section*{References}

\medskip

\small

[1] Asli Celikyilmaz, Antoine Bosselut, Xiaodong He, and Yejin Choi. 2018. Deep communicating agents for abstractive summarization. In Proceedings of the NAACL Conference.

[2] Romain Paulus, Caiming Xiong, and Richard Socher. 2018. A deep reinforced model for abstractive summarization. In Proceedings of the ICLR Conference.

[3] Abigail See, Peter J. Liu, and Christopher D. Manning. 2017. Get to the point: Summarization with pointer-generator networks. In Proceedings of the ACL Conference.

[4] Ashish Vaswani, Noam Shazeer, Niki Parmar, Jakob Uszkoreit, Llion Jones, Aidan N Gomez, Łukasz Kaiser, and Illia Polosukhin. 2017. Attention is all
you need. In Advances in Neural Information Processing Systems, pages 5998–6008.

[5] Caglar Gulcehre, Sungjin Ahn, Ramesh Nallapati, Bowen Zhou, and Yoshua Bengio. 2016. Pointing the unknown words. In Proceedings of the ACL Conference.

[6] Ramesh Nallapati, Bowen Zhou, Cicero dos Santos, Caglar Gulcehre, and Bing Xiang. 2016. Abstractive text summarization using sequence-to-sequence RNNs and beyond. In Computational Natural Language Learning.

[7] Zhengyan Zhang, Xu Han, Zhiyuan Liu, Xin Jiang, Maosong Sun, and Qun Liu. 2019. ERNIE: Enhanced Language Representation with Informative Entities. In Proceedings of the ACL Conference.

[8] Zihang Dai, Zhilin Yang, Yiming Yang, Jaime G. Carbonell, Quoc V. Le, and Ruslan Salakhutdinov. 2019. Transformer-XL: Attentive Language Models beyond a Fixed-Length Context. In Proceedings of the ACL Conference.

[9] Sebastian Gehrmann, Yuntian Deng and Alexander M. Rush. 2018. Bottom-Up Abstractive Summarization. In Proceedings of the EMNLP Conference.

[10] Yang Liu. 2019. Fine-tune BERT for Extractive Summarization. ArXiv abs/1903.10318

[11] Qingyu Zhou, Nan Yang, Furu Wei, Shaohan Huang,
Ming Zhou, and Tiejun Zhao. 2018. Neural document summarization by jointly learning to score and
select sentences. In Proceedings of the ACL Conference.

[12] Jacob Devlin, Ming-Wei Chang, Kenton Lee and Kristina Toutanova. 2018. BERT: Pre-training of Deep Bidirectional Transformers for Language Understanding. In Proceedings of the NAACL Conference.

[13] Karl Moritz Hermann, Tomas Kocisky, Edward Grefenstette, Lasse Espeholt, Will Kay, Mustafa Suleyman, and Phil Blunsom. 2015. Teaching machines to read and comprehend. In Neural Information Processing Systems.

[14] Antoine Bordes, Nicolas Usunier, Alberto García-Durán, Jason Weston and Oksana Yakhnenko. 2013. Translating Embeddings for Modeling Multi-relational Data. In Neural Information Processing Systems.

[15] https://www.wikidata.org/

[16] Chin-Yew Lin. 2004. Rouge: A package for automatic evaluation of summaries. In Text summarization branches out: ACL workshop.

[17] pypi.python.org/pypi/pyrouge/0.1.3

[18] Jianpeng Cheng and Mirella Lapata. 2016. Neural
summarization by extracting sentences and words.
In Proceedings of the ACL Conference.

\end{document}